\documentclass[11pt]{article}
\usepackage{acl2016}
\usepackage{times}
\usepackage{latexsym}
\usepackage{amsmath,amssymb}
\usepackage{amsfonts}
\usepackage{url}
\usepackage{graphicx}
\usepackage{array}
\usepackage{enumitem}

\aclfinalcopy % Uncomment this line for the final submission
\def\aclpaperid{***} %  Enter the acl Paper ID here

\newcolumntype{C}[1]{>{\centering\let\newline\\\arraybackslash\hspace{0pt}}m{#1}}
\title{Joint Learning of the Embedding of Words and Entities for Named Entity Disambiguation}
\author{
  \begin{tabular}{C{3.3cm} C{3.5cm} C{3.1cm} C{4.5cm}}
  Ikuya Yamada$^{1\,3\,4}$ & Hiroyuki Shindo$^2$ & Hideaki Takeda$^3$ & Yoshiyasu Takefuji$^4$ \\
  {\tt \footnotesize{ikuya@ousia.jp}} & {\tt \footnotesize{shindo@is.naist.jp}} & {\tt \footnotesize{takeda@nii.ac.jp}} & {\tt \footnotesize{takefuji@sfc.keio.ac.jp}} \\
  \end{tabular}
  \\
  \\
  {\normalsize $^1$Studio Ousia, 4489-105-221 Endo, Fujisawa, Kanagawa, Japan} \\
  {\normalsize $^2$Nara Institute of Science and Technology, 8916-5 Takayama, Ikoma, Nara, Japan} \\
  {\normalsize $^3$National Institute of Informatics, 2-1-2 Hitotsubashi, Chiyoda, Tokyo, Japan} \\
  {\normalsize $^4$Keio University, 5322 Endo, Fujisawa, Kanagawa, Japan}
}
\date{}
\begin{document}
\maketitle
\begin{abstract}
Named Entity Disambiguation (NED) refers to the task of resolving multiple named entity mentions in a document to their correct references in a knowledge base (KB) (e.g., Wikipedia).
In this paper, we propose a novel embedding method specifically designed for NED.
The proposed method \textit{jointly} maps words and entities into the same continuous vector space.
We extend the \textit{skip-gram} model by using two models.
The \textit{KB graph model} learns the relatedness of entities using the link structure of the KB, whereas the \textit{anchor context model} aims to align vectors such that similar words and entities occur close to one another in the vector space by leveraging KB anchors and their context words.
By combining contexts based on the proposed embedding with standard NED features, we achieved state-of-the-art accuracy of 93.1\% on the standard CoNLL dataset and 85.2\% on the TAC 2010 dataset.
% Our code and pre-trained vectors will be made available online.
\end{abstract}
\section{Introduction}
\label{sec:introduction}

Named Entity Disambiguation (NED) is the task of resolving ambiguous mentions of entities to their referent entities in a knowledge base (KB) (e.g., Wikipedia).
NED has lately been extensively studied \cite{Cucerzan2007,Mihalcea2007,Milne2008,Ratinov2011} and used as a fundamental component in numerous tasks, such as information extraction, knowledge base population \cite{McNamee2009,Ji2010}, and semantic search \cite{Blanco2015a}.
We use Wikipedia as KB in this paper.

The main difficulty in NED is ambiguity in the meaning of entity mentions.
For example, the mention ``Washington'' in a document can refer to various entities, such as the state, or the capital of the US, the actor \textsf{Denzel Washington}, the first US president \textsf{George Washington}, and so on.
In order to resolve these ambiguous mentions into references to the correct entities, early approaches focused on modeling \textit{textual} context, such as the similarity between contextual words and encyclopedic descriptions of a candidate entity \cite{Bunescu2006,Mihalcea2007}.
Most state-of-the-art methods use more sophisticated \textit{global} approaches, where all mentions in a document are simultaneously disambiguated based on global \textit{coherence} among disambiguation decisions.

Word embedding methods are also becoming increasingly popular \cite{Mikolov2013,Mikolov2013a,Pennington2014}.
These involve learning continuous vector representations of words from large, unstructured text corpora.
The vectors are designed to capture the semantic similarity of words when similar words are placed near one another in a relatively low-dimensional vector space.

In this paper, we propose a method to construct a novel embedding that \textit{jointly} maps words and entities into the same continuous vector space.
In this model, similar words and entities are placed close to one another in a vector space.
Hence, we can measure the similarity between any pair of items (i.e., words, entities, and a word and an entity) by simply computing their cosine similarity.
This enables us to easily measure the contextual information for NED, such as the similarity between a context word and a candidate entity, and the relatedness of entities required to model coherence.

Our model is based on the skip-gram model \cite{Mikolov2013,Mikolov2013a}, a recently proposed embedding model that learns to predict each context word given the target word.
Our model consists of the following three models based on the skip-gram model:
1) the conventional skip-gram model that learns to predict neighboring words given the target word in text corpora,
2) the \textit{KB graph model} that learns to estimate neighboring entities given the target entity in the link graph of the KB,
and 3) the \textit{anchor context model} that learns to predict neighboring words given the target entity using anchors and their context words in the KB.
By jointly optimizing these models, our method simultaneously learns the embedding of words and entities.

Based on our proposed embedding, we also develop a straightforward NED method that computes two contexts using the proposed embedding: textual context similarity, and coherence.
Textual context similarity is measured according to vector similarity between an entity and words in a document.
Coherence is measured based on the relatedness between the target entity and other entities in a document.
Our NED method combines these contexts with several standard features (e.g., prior probability) using supervised machine learning.

We tested the proposed method using two standard NED datasets: the CoNLL dataset and the TAC 2010 dataset.
Experimental results revealed that our method outperforms state-of-the-art methods on both datasets by significant margins.
Moreover, we conducted experiments to separately assess the quality of the vector representation of entities using an entity relatedness dataset, and discovered that our method successfully learns the quality representations of entities.

\section{Joint Embedding of Words and Entities}

In this section, we first describe the conventional skip-gram model for learning word embedding.
We then explain our method to construct an embedding that jointly maps words and entities into the same continuous $d$-dimensional vector space.
We extend the skip-gram model by adding the \textit{KB graph model} and the \textit{anchor context model}.

\subsection{Skip-gram Model for Word Similarity}
\label{subsec:skip-gram-model}

The training objective of the skip-gram model is to find word representations that are useful to predict context words given the target word.
Formally, given a sequence of $T$ words $w_1, w_2, ..., w_T$, the model aims to maximize the following objective function:
\begin{equation}
\mathcal{L}_w = \sum_{t=1}^{T}\sum_{-c \leq j \leq c,j \neq 0}\log P(w_{t+j}|w_t)
\label{objective_skipgram}
\end{equation}
where $c$ is the size of the context window, $w_t$ denotes the target word, and $w_{t+j}$ is its context word.
The conditional probability $P(w_{t+j}|w_t)$ is computed using the following softmax function:
\begin{equation}
P(w_{t+j}|w_t) = \frac{\exp(\mathbf{V}_{w_t}\!^\top \mathbf{U}_{w_{t+j}})}{\sum_{w \in W}\exp(\mathbf{V}_{w_t}\!^\top \mathbf{U}_w)}
\label{softmax_skipgram}
\end{equation}
where $W$ is a set containing all words in the vocabulary, and $\mathbf{V}_w \in \mathbb{R}^d$ and $\mathbf{U}_w \in \mathbb{R}^d$ denote the vectors of word $w$ in matrices $\mathbf{V}$ and $\mathbf{U}$, respectively.

The skip-gram model is trained to optimize the above function $\mathcal{L}_w$, and $\mathbf{V}$ are used as the resulting vector representations of words.

\subsection{Extending the Skip-gram Model}

We extend the skip-gram model to learn the vector representations of entities.
We expand matrices $\textbf{V}$ and $\textbf{U}$ to include the vectors of entities $\mathbf{V}_e \in \mathbb{R}^d$ and $\mathbf{U}_e \in \mathbb{R}^d$ in addition to the vectors for words.

\subsubsection{KB Graph Model}

We use an internal link structure in KB to enable the model to learn the relatedness between pairs of entities.
Wikipedia Link-based Measure (WLM) \cite{DavidMilne} is a method to measure entity relatedness based on its link structure.
It has been used as a standard method to compute the relatedness of entities for modeling coherence in past NED studies.
The relatedness between two entities is computed using the following function:
\begin{equation}
\resizebox{.89\hsize}{!}{$
WLM(e_1, e_2) = 1 - \frac{\log\,\max(|C_{e_1}|, |C_{e_2}|) - \log|C_{e_1} \cap C_{e_2}|}{\log|E| - \log\,\min(|C_{e_1}|, |C_{e_2}|)}
$}
\label{wlm}
\end{equation}
where $E$ is the set of all entities in KB and $C_e$ is the set of entities with a link to an entity $e$.
Intuitively, WLM assumes that entities with similar incoming links are related.
Despite its simplicity, WLM yields state-of-the-art performance \cite{Hoffart2012}.

Inspired by WLM, the KB graph model simply learns to place entities with similar incoming links near one another in the vector space.
We formalize this as the following objective function:
\begin{equation}
\mathcal{L}_e = \sum_{e_i \in E}\sum_{e_o \in C_{e_i}, e_i \neq e_o}\log P(e_o|e_i)
\label{objective_entity_rel}
\end{equation}
We compute the conditional probability $P(e_o|e_i)$ using the following softmax function:
\begin{equation}
P(e_o|e_i) = \frac{\exp(\mathbf{V}_{e_i}\!^\top \mathbf{U}_{e_o})}{\sum_{e \in E}\exp(\mathbf{V}_{e_i}\!^\top \mathbf{U}_e)}
\label{softmax_entity_rel}
\end{equation}
We train the model to predict the incoming links $C_{e}$ given an entity $e$.
Therefore, $C_{e}$ plays a similar role to context words in the skip-gram model.

\subsubsection{Anchor Context Model}

If we add only the KB graph model to the skip-gram model, the vectors of words and entities do not interact, and can be placed in different subspaces of the vector space.
To address this issue, we introduce the anchor context model to place similar words and entities near one another in the vector space.

The idea underlying this model is to leverage KB anchors and their context words to train the model.
As mentioned in Section \ref{sec:introduction}, we use Wikipedia as a KB.
It contains many internal anchors that can be safely treated as unambiguous occurrences of referent KB entities.
By using these anchors, we can easily obtain many occurrences of entities and their corresponding context words directly from the KB.

As in the skip-gram model, we simply train the model to predict the context words of an entity pointed to by the target anchor.
The objective function is as follows:
\begin{equation}
\mathcal{L}_a = \sum_{(e_i, Q) \in A}\sum_{w_o \in Q}\log P(w_o|e_i)
\label{objective_alignment}
\end{equation}
where $A$ denotes a set of anchors in the KB, each of which contains a pair of a referent entity $e_i$ and a set of its context words $Q$.
Here, $Q$ contains the previous $c$ words and the next $c$ words.
Note that $|A|$ equals the number of internal anchors in the KB.
As in past models, the conditional probability $P(w_o|e_i)$ is computed using the softmax function:
\begin{equation}
P(w_o|e_i) = \frac{\exp(\mathbf{V}_{e_i}\!^\top \mathbf{U}_{w_o})}{\sum_{w \in W}\exp(\mathbf{V}_{e_i}\!^\top \mathbf{U}_w)}
\label{softmax_alignment}
\end{equation}
Using the proposed model, we align the vector representations of words and entities by placing words and entities with similar context words close to one another in the vector space.

\subsection{Training}

Considering the three model components mentioned above, we propose the following objective function by linearly combining the above objective functions:
\begin{equation}
\mathcal{L} = \mathcal{L}_w + \mathcal{L}_e + \mathcal{L}_a
\end{equation}
The training of the model is intended to maximize the above function, and the resulting matrix $\mathbf{V}$ is used to embed words and entities.

One of the problems in training our model is that the normalizers contained in the softmax functions $P(w_{t+j}|w_t)$, $P(e_o|e_i)$, and $P(w_o|e_i)$ are computationally very expensive because they involve summation over all words $W$ or entities $E$.
To address this problem, we use \textit{negative sampling (NEG)} \cite{Mikolov2013a} to convert original objective functions into computationally feasible ones.
NEG is defined by the following objective function:
\begin{equation}
\resizebox{.89\hsize}{!}{$
\log\sigma(\mathbf{V}_{w_t} \!^\top \mathbf{U}_{w_{t+j}}) + \sum_{i=1}^{g}\mathbb{E}_{w_i \sim P_{neg}(w)} \Big[\log\sigma(-\mathbf{V}_{w_t} \!^\top \mathbf{U}_{w_i})\Big]
$}
\end{equation}
where $\sigma(x) = 1/(1 + \exp(-x))$ and $g$ is the number of negative samples.
We replace the $\log P(w_{t+j}|w_t)$ term in Eq. \eqref{objective_skipgram} with the above objective function.
Consequently, the objective function is transformed from that in Eq. \eqref{objective_skipgram} to a simple objective function of the binary classification to distinguish the observed word $w_t$ from words drawn from noise distribution $P_{neg}(w)$.
We also replace $\log P(e_o|e_i)$ in Eq. \eqref{objective_entity_rel} and $\log P(w_o|e_i)$ in Eq. \eqref{objective_alignment} in the same manner.

Note that NEG takes a negative distribution $P_{neg}(w)$ as a free parameter.
Following \cite{Mikolov2013a}, we use the unigram distribution of words ($U(w)$) raised to the $3/4^{th}$ power (i.e., $U(w)^{3/4}/Z$, where $Z$ is a normalization constant) in the skip-gram model and the anchor context model.
In the KB graph model, we use a uniform distribution over KB entities $E$ as the negative distribution.

We use Wikipedia to train all the above models.
Optimization is carried out simultaneously to maximize the transformed objective function by iterating over Wikipedia pages several times.
We use stochastic gradient descent (SGD) for the optimization.
The optimization is performed using a multiprocess-based implementation of our model using Python, Cython, and NumPy configured with OpenBLAS with storing matrices $\mathbf{V}$ and $\mathbf{U}$ in the shared memory.
To improve speed, we decide not to introduce locks to the shared matrices.

\section{Named Entity Disambiguation Using Embedding}

In this section, we explain our NED method using our proposed embedding.
Let us formally define the task.
Given a set of entity mentions $M = \{m_1, m_2, ..., m_N\}$ in a document $d$ with an entity set $E = \{e_1, e_2, ..., e_K\}$ in the KB, the task is defined as resolving mentions (e.g., ``Washington'') into their referent entities (e.g., \textsf{Washington D.C.}).

We introduce two measures that have been frequently observed in past NED studies: \textit{entity prior} $P(e)$ and \textit{prior probability} $P(e|m)$.
We define entity prior $P(e) = |A_{e,*}| / |A_{*,*}|$ where $A_{*,*}$ denotes all anchors in the KB and $A_{e,*}$ is the set of anchors that point to entity $e$.
Prior probability is defined as $P(e|m) = |A_{e,m}| / |A_{*,m}|$ where $A_{*,m}$ represents all anchors with the same surface as mention $m$ in KB and $A_{e, m}$ is a subset of $A_{*,m}$ that points to entity $e$.

We separate the NED task into two sub-tasks: \textit{candidate generation} and \textit{mention disambiguation}.
In candidate generation, candidates of referent entities are generated for each mention.
Details of candidate generation are provided in Section \ref{subsubsec:ned-setup}.

\subsection{Mention Disambiguation}

Given a document $d$ and mention $m$ with its candidate referent entities $\{e_1, e_2, ..., e_k\}$ generated in the candidate generation step, the task is to disambiguate mention $m$ by selecting the most relevant entity from the candidate entities.

The key to improving the performance of this task is to effectively model the context.
We propose two novel methods to model the context using the proposed embedding.
Further, we combine these two models with several standard NED features using supervised machine learning described in \ref{subsubsec:learning-to-rank}.

\subsubsection{Modeling Textual Context}
\label{subsubsec:textual-context}

Textual context is designed based on the assumption that an entity is more likely to appear if the context of a given mention is similar to that of the entity.

We propose a method to measure the similarity between textual context and entity using the proposed embedding by first deriving the vector representation of the context and then computing the similarity between the context and the entity using cosine similarity.
To derive the vector of context, we average the vectors of context words:
\begin{equation}
\vec{v_{c_w}} = \frac{1}{|W_{c_m}|}\sum_{w \in W_{c_m}}\vec{v_{w}}
\label{eq:word-vector-average}
\end{equation}
where $W_{c_m}$ is a set of the context words of mention $m$ and $\vec{v_w} \in \mathbf{V}$ denotes the vector representation of word $w$.
We use all noun words in document $d$ as context words.\footnote{We used Apache OpenNLP tagger to detect nouns. {\url{https://opennlp.apache.org/}}}
Moreover, we ignore a context word if the surface of mention $m$ contains it.

We then measure the similarity between candidate entity and the derived textual context by using cosine similarity between $\vec{v_{c_w}}$ and the vector of entity $\vec{v_e}$.

\subsubsection{Modeling Coherence}
\label{subsubsec:coherence}

It has been revealed that effectively modeling coherence in the assignment of entities to mentions is important for NED.
However, this is a chicken-and-egg problem because the assignment of entities to mentions, which is required to measure coherence, is not possible prior to performing NED.

% To address this problem, we introduce a simple \textit{two-step} approach:
Similar to past work \cite{Ratinov2011}, we address this problem by employing a simple \textit{two-step} approach:
we first train the machine learning model using the coherence score among unambiguous mentions\footnote{We consider that mention $m$ unambiguously refers to entity $e$ if its prior probability $P(e|m)$ is greater than 0.95.}, in addition to other features, and then retrain the model using the coherence score among the predicted entity assignments instead.

To estimate coherence, we first calculate the vector representation of the context entities and measure the similarity between the vector of the context entities and that of the target entity $e$.
Note that context entities are unambiguous entities in the first step, and predicted entities are used instead in the second step.

To derive the vector representation of context entities, we average their vector representations:
\begin{equation}
\vec{v_{c_e}} = \frac{1}{|E_{c_m}|}\sum_{e^* \in E_{c_m}}\vec{v_{e^*}}
\end{equation}
where $E_{c_m}$ denotes the set of context entities described above.

To estimate the coherence score, we again use cosine similarity between the vector of entity $\vec{v}_e$ and that of context entities $\vec{v}_{c_e}$.

\subsubsection{Learning to Rank}
\label{subsubsec:learning-to-rank}

To combine the proposed contextual information described above with standard NED features, we employ a method of supervised machine learning to rank the candidate entities given mention $m$ and document $d$.

In particular, we use Gradient Boosted Regression Trees (GBRT) \cite{Friedman2001}, a state-of-the-art point-wise learning-to-rank algorithm widely used for various tasks, which has been recently adopted for the sort of tasks for which we employ it here \cite{Meij2012}.
GBRT consists of an ensemble of regression trees, and predicts a relevance score given an instance.
We use the GBRT implementation in \textit{scikit-learn}\footnote{\url{http://scikit-learn.org/}} and the logistic loss is used as the loss function.
The main parameters of GBRT are the number of iterations $\eta$, the learning rate $\beta$, and the maximum depth of the decision trees $\xi$.

With regard to the features of machine learning, we first use prior probability ($P(e|m)$) and entity prior ($P(e)$).
Further, we include a feature representing the maximum prior probability of the candidate entity $e$ of all mentions in the document.
We also add the number of entity candidates for mention $m$ as a feature.
The above set of four features is called \textit{base} features in the rest of the paper.

We also use several \textit{string similarity} features used in past work on NED \cite{Meij2012}.
These features aim to capture the similarity between the title of entity $e$ and the surface of mention $m$, and consist of the edit distance, whether the title of entity $e$ exactly equals or contains the surface of mention $m$, and whether the title of entity $e$ starts or ends with the surface of mention $m$.

Finally, we include contextual features measured using the proposed embedding.
We use cosine similarity between the candidate entity and the textual context (see Section \ref{subsubsec:textual-context}), and similarity between an entity and contextual entities (see Section \ref{subsubsec:coherence}).
Furthermore, we include the rank of entity $e$ among candidate entities of mention $m$, sorted according to these two similarity scores in descending order.

\section{Experiments}
\label{sec:experiments}

% \begin{table}[t]
% \centering
% \begin{tabular}{l|ccc}
% \hline
% & WordSim-353 & MC & RG \\
% \hline
% Our Method & 0.66 & \textbf{0.78} & \textbf{0.77} \\
% % python scripts/word_similarity/evaluate_wordsim_353.py dataset/wordsim353/combined.tab enwiki_entity_vector_500_20150923_10_0.05_10
% % Pearson score: 0.6455
% % Spearman score: 0.6639
% % python scripts/word_similarity/evaluate_mc_30.py dataset/MC-30/EN-MC-30.txt enwiki_entity_vector_500_20150923_10_0.05_10
% % Pearson score: 0.7945
% % Spearman score: 0.7759
% % python scripts/word_similarity/evaluate_rg_65.py dataset/RG-65/EN-RG-65.txt enwiki_entity_vector_500_20150923_10_0.05_10
% % Pearson score: 0.7675
% % Spearman score: 0.7704
% Skip-gram & \textbf{0.67} & 0.77 & 0.76 \\
% \hline
% \end{tabular}
% \caption{Results of the word similarity task.}
% \label{tb:word-sim-scores}
% \end{table}

In this section, we describe the setup and results of our experiments.
In addition to experiments on the NED task, we separately assessed the quality of pairwise \textit{entity relatedness} in order to test the effectiveness of our method in capturing pairwise similarity between pairs of entities.
We first describe the details of the training of the embedding and then present the experimental results.

\subsection{Training for the Proposed Embedding}

To train the proposed embedding, we used the December 2014 version of the Wikipedia dump\footnote{The dump was retrieved from Wikimedia Downloads. \url{http://dumps.wikimedia.org/}}.
We first removed the pages for navigation, maintenance, and discussion, and used the remaining 4.9 million pages.
% In [1]: from entity_vector.wiki_dump_reader import WikiDumpReader
% In [2]: dump_reader = WikiDumpReader('enwiki-latest-pages-articles.xml.bz2'
% In [3]: sum([1 for page in dump_reader if not page.is_redirect])
% Out[3]: 4884901
We parsed the Wikipedia pages and extracted text and anchors from each page.
We further tokenized the text using the \textit{Apache OpenNLP} tokenizer.
We also filtered out rare words that appeared fewer than five times in the corpus.
We thus obtained approximately 2 billion tokens and 73 million anchors.
% In [1]: from entity_vector import Dictionary
% In [3]: dictionary = Dictionary.load(open('enwiki_dictionary_20150706.pickle'))
% In [4]: sum([w.count for w in dictionary.words()])
% Out[4]: 1993281946
% In [5]: sum([e.count for e in dictionary.entities()])
% Out[5]: 73551941
% In [7]: len(list(dictionary.words()))
% Out[7]: 2107060
% In [9]: from entity_vector import EntityVector
% In [10]: entity_vector = EntityVector.load('enwiki_entity_vector_300_20150712_8_0.05')
% In [11]: entity_vector.syn0.shape
%Out[11]: (7149787, 300)
The total number of words and entities in the embedding were approximately 2.1 million and 5 million, respectively.
Consequently, the number of rows of matrices $\mathbf{V}$ and $\mathbf{U}$ were 7.1 million.

The number of dimensions $d$ of the embedding was set to 500.
Following \cite{Mikolov2013a}, we also used learning rate $\alpha = 0.025$ which linearly decreased with the iterations of the Wikipedia dump.
Regarding the other parameters, we set the size of the context window $c = 10$ and the negative samples $g = 30$.
The model was trained online by iterating over pages in the Wikipedia dump 10 times.
The training lasted approximately five days using a server with a 40-core CPU on Amazon EC2.

% \subsection{Word Similarity}

% In order to test the quality of vector representations of words, we used three standard word similarity datasets: the \textit{WordSim-353} dataset \cite{Finkelstein2002}, the \textit{MC} dataset \cite{Miller1991}, and the \textit{RG} dataset \cite{Rubenstein1965} that contain 353, 65, and 30 word pairs, respectively.
% Each word pair has a \textit{gold-standard} similarity score assigned by human judges.

% We used cosine similarity to calculate similarity score between any pair of words.
% Following past work, we computed the correlation between similarity scores through human judgments on a set of word pairs using Spearman's rank correlation coefficient.
% Here, we adopted the skip-gram model as baseline.

% We used our implementation to train the skip-gram model.
% Furthermore, the following parameters were used to train the model: $d = 500$, $c = 10$, $g = 30$, and $\alpha = 0.025$.
% We trained the model by iterating over the Wikipedia dump 10 times.

% Table \ref{tb:word-sim-scores} shows the results.
% Compared to the skip-gram model, our method performed comparably on the WordSim-353 dataset, and slightly better on other datasets, thus showing that the proposed extension to the skip-gram model can be also beneficial for improving word representations.

\subsection{Entity Relatedness}
\begin{table}[t]
\centering
\begin{tabular}{l|cccccc}
\hline
& \scriptsize{NDCG@1} & \scriptsize{NDCG@5} & \scriptsize{NDCG@10} & \scriptsize{MAP} \\
\hline
\footnotesize{Our Method} & \textbf{0.59} & \textbf{0.56} & \textbf{0.59} & \textbf{0.52} \\
\footnotesize{WLM} & 0.54 & 0.52 & 0.55 & 0.48 \\
% python evaluate_dexter.py queries.json entity-disambi/enwiki_entity_vector_500_20151026
% NDCG@1: 0.592
% NDCG@5: 0.561
% NDCG@10: 0.587
% MAP: 0.524
% MRR: 0.702
% P@1: 0.592
% P@5: 0.395
% P@10: 0.276
\hline
\end{tabular}
\caption{Results of the entity relatedness task.}
\label{tb:entity-rel-scores}
\end{table}
To test the quality of the vector representation of entities, we conducted an experiment using a dataset for entity relatedness created by Ceccarelli et al. \cite{Ceccarelli2013}.
The dataset consists of training, test, and validation sets, and we only use the test set.
The test set contains 3,314 entities, where each entity has 91 candidate entities with \textit{gold-standard} labels indicating whether the two entities are related.
Following \cite{DBLP:journals/corr/HuangHJ15}, we obtained the ranked order of the candidate entities using cosine similarity between the target entity and each of the candidate entities, and computed the two standard measures: normalized discounted cumulative gain (NDCG) \cite{Jarvelin2002} and mean average precision (MAP) \cite{Manning2008}.
We adopted WLM as baseline.

Table \ref{tb:entity-rel-scores} shows the results.
The score for WLM was obtained from Huang et al. \cite{DBLP:journals/corr/HuangHJ15}.
Our method clearly outperformed WLM.
The results show that our method accurately captures pairwise entity relatedness.

\subsection{Named Entity Disambiguation}

\subsubsection{Setup}
\label{subsubsec:ned-setup}

We now explain our experimental setup for the NED task.
We tested the performance of our proposed method on two standard NED datasets: the \textit{CoNLL} dataset and the \textit{TAC 2010} dataset.
The details of these datasets are provided below.
Moreover, as with the corpus used in the embedding, we used the December 2014 version of the Wikipedia dump as the referent KB, and to derive the prior probability as well as the entity prior.

To find the best parameters for our machine learning model, we ran a parameter search on the CoNLL development set.
We used $\eta = 10,000$ trees, and tested all combinations of the learning rate $\beta = \{0.01, 0.02, 0.03, 0.05\}$ and the maximum depth of the decision trees $\xi = \{3, 4, 5\}$.
We computed their accuracy on the dataset, and found that the parameters did not significantly affect performance (1.0\% at most).
% We computed their accuracy on the CoNLL development set, and found that the parameters did not significantly affect performance (1.0\% at most).
% python scripts/ml_builder/eval_gbrt.py aida_ppr_dataset_testa_0.05_500_2.pickle aida_ppr_gbrt_0.05_500_2
% aida_ppr_gbrt_0.05_500_2_3_0.01_10000.pickle    0.92862
% aida_ppr_gbrt_0.05_500_2_3_0.02_10000.pickle    0.93237
% aida_ppr_gbrt_0.05_500_2_3_0.03_10000.pickle    0.93300
% aida_ppr_gbrt_0.05_500_2_3_0.05_10000.pickle    0.93863
% aida_ppr_gbrt_0.05_500_2_4_0.01_10000.pickle    0.93237
% aida_ppr_gbrt_0.05_500_2_4_0.02_10000.pickle    0.93738
% aida_ppr_gbrt_0.05_500_2_4_0.03_10000.pickle    0.93697
% aida_ppr_gbrt_0.05_500_2_4_0.05_10000.pickle    0.93780
% aida_ppr_gbrt_0.05_500_2_5_0.01_10000.pickle    0.93884
% aida_ppr_gbrt_0.05_500_2_5_0.02_10000.pickle    0.93884
% aida_ppr_gbrt_0.05_500_2_5_0.03_10000.pickle    0.93697
% aida_ppr_gbrt_0.05_500_2_5_0.05_10000.pickle    0.93550
We used $\beta = 0.02$ and $\xi = 4$ which yielded the best performance.

\paragraph*{CoNLL}

The CoNLL dataset is a popular NED dataset constructed by Hoffart et al. \cite{Hoffart2011}.
The dataset is based on NER data from the CoNLL 2003 shared task, and consists of training, development, and test sets, containing 946, 216, and 231 documents, respectively.
We trained our machine learning model using the training set and reported its performance using the test set.
We also used the development set for the parameter tuning described above.
Following \cite{Hoffart2011}, we only used 27,816 mentions with valid entries in the KB and reported the standard micro- (aggregates over all mentions) and macro- (aggregates over all documents) accuracies of the top-ranked candidate entities to assess disambiguation performance.
For candidate generation, we use the following two resources: 1) a public dataset recently built by Pershina et al. \cite{pershina-he-grishman:2015:NAACL-HLT} (denoted by \textit{PPRforNED}) for the sake of compatibility with their state-of-the-art results, and 2) a dictionary built using a standard YAGO means relation dataset \cite{Hoffart2011} (denoted by \textit{YAGO}).
Moreover, we used PPRforNED for the parameter tuning of the machine learning model and for error analysis.

\paragraph*{TAC 2010}

The TAC 2010 dataset is another popular NED dataset constructed for the Text Analysis Conference (TAC)\footnote{\url{http://www.nist.gov/tac/}} \cite{Ji2010}.
The dataset is based on news articles from various agencies and Web log data, and consists of a training and a test set containing 1,043 and 1,013 documents, respectively.
Following past work \cite{he-EtAl:2013:Short,TACL494}, we used mentions only with a valid entry in the KB, and reported the micro-accuracy score of the top-ranked candidate entities.
We trained our model using the training set and assessed its performance using the test set.
Consequently, we evaluated our model on 1,020 mentions contained in the test set.
For candidate generation, we used a dictionary that was directly built from the Wikipedia dump mentioned previously.
Similar to past work, we retrieved possible mention surfaces of an entity from (1) the title of the entity, (2) the title of another entity redirecting to the entity, and (3) the names of anchors that point to the entity.
% Following Cucerzan \cite{Cucerzan2007}, these titles are normalized by removing trailing parentheses\footnote{For example, the entity title \textit{Texas (TV series)} was normalized to \textit{Texas}.}.
% Further, we tokenized each entity's title and included each word as a candidate surface.
We retained the top 50 candidates through their entity priors for computational efficiency.

\subsubsection{Comparison with State-of-the-art Methods}

We compared our method with the following recently proposed state-of-the-art methods:
\begin{itemize}[itemsep=0em,topsep=0.3em]
\item Hoffart et al. \cite{Hoffart2011} is a graph-based approach that finds a dense subgraph of entities in a document to address NED.
\item He et al. \cite{he-EtAl:2013:Short} uses deep neural networks to derive the representations of entities and mention contexts and applies them to NED.
\item Chisholm and Hachey \cite{TACL494} uses a Wikilinks dataset \cite{singh12:wiki-links} to improve the performance of NED.
\item Pershina et al. \cite{pershina-he-grishman:2015:NAACL-HLT} improved NED by modeling coherence using the personalized page rank algorithm, and achieved the best-known accuracy on the CoNLL dataset.
% \item \textit{LCC} \cite{Lehmann2010} was a top-ranked system in the TAC 2010 competition.
% \item \textit{AIDA} \cite{Hoffart2011} is a graph-based approach that finds a dense subgraph of entities in a document to address NED.
% \item \textit{He} \cite{he-EtAl:2013:Short} uses deep neural networks to derive the representations of entities and mention contexts and applies them to NED.
% \item \textit{Alhelbawy} \cite{alhelbawy-gaizauskas:2014:P14-2} adopts the page rank algorithm to address NED.
% \item \textit{Huang} \cite{DBLP:journals/corr/HuangHJ15} adopts deep neural networks to learn entity representations such that its pairwise relatedness is suitable for NED, and uses a graph-based algorithm to address NED.
% \item \cite{Sum2015} uses deep neural networks to model the representations of mentions, mention contexts, and entities and applies them to NED.
\end{itemize}
\subsubsection{Results}

\begin{table}[tb]
\centering
\begin{tabular}{l|cc}
\hline
& \begin{tabular}{@{}c@{}}Micro\\accuracy\end{tabular} & \begin{tabular}{@{}c@{}}Macro\\accuracy\end{tabular} \\
\hline
CoNLL (PPRforNED) & 93.1 & 92.6 \\
CoNLL (YAGO) & 91.5 & 90.9 \\
TAC 2010 & 85.2 & - \\
\hline
\end{tabular}
\caption{Experimental results of our proposed NED method.}

\label{tb:ned-results}
\end{table}

\begin{table}[t]
\centering
\begin{tabular}{l|ccc}
\hline
& \begin{tabular}{@{}c@{}} \footnotesize{CoNLL} \\ \footnotesize{(Micro)}\end{tabular} & \begin{tabular}{@{}c@{}}\footnotesize{CoNLL} \\ \footnotesize{(Macro)}\end{tabular} & \begin{tabular}{@{}c@{}} \footnotesize{TAC10} \\ \footnotesize{(Micro)}\end{tabular} \\
% & \begin{tabular}{@{}c@{}} \footnotesize{CoNLL} \\ \footnotesize{(Micro)}\end{tabular} & \begin{tabular}{@{}c@{}}\footnotesize{CoNLL} \\ \footnotesize{(Macro)}\end{tabular} & \footnotesize{TAC10} \\
\hline
\footnotesize{Our Method} & \textbf{93.1} & \textbf{92.6} & \textbf{85.2} \\
\hline
\footnotesize{Hoffart et al., 2011} & 82.5 & 81.7 & - \\
\footnotesize{He et al., 2013} & 85.6 & 84.0 & 81.0 \\
\footnotesize{Chisholm \& Hachey, 2015} & 88.7 & - & 80.7 \\
\footnotesize{Pershina et al., 2015} & 91.8 & 89.9 & - \\
\hline
\end{tabular}
\caption{Accuracy scores of the proposed method and the state-of-the-art methods.}
\label{tb:state-of-the-art}
\end{table}

Table \ref{tb:ned-results} shows the experimental results of our proposed method.
Our method successfully achieved enhanced performance on both the CoNLL and the TAC 2010 datasets.
Moreover, we found that the choice of candidate generation method considerably affected performance on the CoNLL dataset.

Further, Table \ref{tb:state-of-the-art} shows the experimental results of our proposed method as well as those of state-of-the-art methods.
Our method outperformed all the state-of-the-art methods on both datasets.
% Note that because Pershina et al. uses the whole CoNLL dataset rather than the test set to evaluate their performance, the results cannot be directly comparable to others.

\subsubsection{Feature Study}

\begin{table}[tb]
\centering
\begin{tabular}{p{3.8cm}|cc}
\hline
& \begin{tabular}{@{}c@{}}Micro\\accuracy\end{tabular} & \begin{tabular}{@{}c@{}}Macro\\accuracy\end{tabular} \\
\hline
\textbf{CoNLL (PPRforNED):}\\
\hline
Base & 85.4 & 87.4 \\
% time venv; entity-disambi evaluation --corpus-type=aida_ppr --corpus-tag=testb dataset/aida-ppr enwiki_dictionary_20150706.pickle enwiki_alias_db_20150928.pickle ml aida_ppr_gbrt_0.05_500_base_4_0.02_10000.pickle enwiki_entity_vector_500_20150923_10_0.05_10
% Processing time: 57.994
% Average time per document: 0.251
% Correct mentions: 3829
% Total mentions: 4485
% Precision (Micro): 0.8537
% Precision (Macro): 0.8737
+String similarity & 85.8 & 87.8 \\
% venv; entity-disambi evaluation --corpus-type=aida_ppr --corpus-tag=testb dataset/aida-ppr enwiki_dictionary_20150706.pickle enwiki_alias_db_20150928.pickle ml aida_ppr_gbrt_0.05_500_strsim_4_0.02_10000.pickle enwiki_entity_vector_500_20150923_10_0.05_10
% Processing time: 86.283
% Average time per document: 0.374
% Correct mentions: 3847
% Total mentions: 4485
% Precision (Micro): 0.8577
% Precision (Macro): 0.8778
+Textual context & 90.9 & 92.4 \\
% time venv; entity-disambi evaluation --corpus-type=aida_ppr --corpus-tag=testb dataset/aida-ppr enwiki_dictionary_20150706.pickle enwiki_alias_db_20150928.pickle ml aida_ppr_gbrt_0.05_500_txt_4_0.02_10000.pickle enwiki_entity_vector_500_20150923_10_0.05_10
% Processing time: 56.068
% Average time per document: 0.243
% Correct mentions: 4076
% Total mentions: 4485
% Precision (Micro): 0.9088
% Precision (Macro): 0.9241
+Coherence & 91.4 & 92.1 \\
% time venv; entity-disambi evaluation --corpus-type=aida_ppr --corpus-tag=testb dataset/aida-ppr enwiki_dictionary_20150706.pickle enwiki_alias_db_20150928.pickle ml aida_ppr_gbrt_0.05_500_4_0.02_10000.pickle enwiki_entity_vector_500_20150923_10_0.05_10
% Processing time: 116.139
% Average time per document: 0.503
% Correct mentions: 4101
% Total mentions: 4485
% Precision (Micro): 0.9144
% Precision (Macro): 0.9210
Two-step & \textbf{93.1} & \textbf{92.6} \\
% time venv; entity-disambi evaluation --corpus-type=aida_ppr --corpus-tag=testb dataset/aida-ppr enwiki_dictionary_20150706.pickle enwiki_alias_db_20150928.pickle ml aida_ppr_two_step_gbrt_0.05_500_4_0.02_10000.pickle enwiki_entity_vector_500_20150923_10_0.05_10
% Processing time: 177.757
% Average time per document: 0.770
% Correct mentions: 4174
% Total mentions: 4485
% Precision (Micro): 0.9307
% Precision (Macro): 0.9259
\hline
\textbf{CoNLL (YAGO):}\\
\hline
Base & 81.1 & 83.6 \\
% % venv; entity-disambi evaluation --corpus-type=aida --corpus-tag=testb dataset/aida-yago2 enwiki_dictionary_20150706.pickle yago_alias_db_20150819.pickle ml aida_gbrt_10000_0.02_4_base.pickle enwiki_entity_vector_500_20150923_10_0.05_10
% Correct mentions: 3638
% Total mentions: 4484
% Precision (Micro): 0.8113
% Precision (Macro): 0.8359
+String similarity & 81.3 & 84.2 \\
% % venv; entity-disambi evaluation --corpus-type=aida --corpus-tag=testb dataset/aida-yago2 enwiki_dictionary_20150706.pickle yago_alias_db_20150819.pickle ml aida_gbrt_10000_0.02_4_strsim.pickle enwiki_entity_vector_500_20150923_10_0.05_10
% Correct mentions: 3645
% Total mentions: 4484
% Precision (Micro): 0.8129
% Precision (Macro): 0.8417
+Textual context & 87.2 & 89.6 \\
% % venv; entity-disambi evaluation --corpus-type=aida --corpus-tag=testb dataset/aida-yago2 enwiki_dictionary_20150706.pickle yago_alias_db_20150819.pickle ml aida_gbrt_10000_0.02_4_txt.pickle enwiki_entity_vector_500_20150923_10_0.05_10
% Correct mentions: 3912
% Total mentions: 4484
% Precision (Micro): 0.8724
% Precision (Macro): 0.8955
+Coherence & 90.3 & 90.8 \\
% % venv; entity-disambi evaluation --corpus-type=aida --corpus-tag=testb dataset/aida-yago2 enwiki_dictionary_20150706.pickle yago_alias_db_20150819.pickle ml aida_gbrt_10000_0.02_4.pickle enwiki_entity_vector_500_20150923_10_0.05_10
% Correct mentions: 4048
% Total mentions: 4484
% Precision (Micro): 0.9028
% Precision (Macro): 0.9081
Two-step & \textbf{91.5} & \textbf{90.9} \\
\hline
\textbf{TAC 2010:} \\
\hline
Base & 80.1 & - \\
% venv; entity-disambi evaluation --corpus-type=tac_kbp --corpus-tag=eval dataset/tac-kbp-2010/ enwiki_dictionary_20150706.pickle enwiki_alias_db_20150927_tt.pickle ml tac_gbrt_0.05_500_base_4_0.02_10000.pickle enwiki_entity_vector_500_20150923_10_0.05_10
% Processing time: 144.016
% Average time per document: 0.142
% Correct mentions: 826
% Total mentions: 1020
% Precision (Micro): 0.8098
% Precision (Macro): 0.8091
+String similarity & 81.7 & - \\
% venv; entity-disambi evaluation --corpus-type=tac_kbp --corpus-tag=eval dataset/tac-kbp-2010/ enwiki_dictionary_20150706.pickle enwiki_alias_db_20150927_tt.pickle ml tac_gbrt_0.05_500_strsim_4_0.02_10000.pickle enwiki_entity_vector_500_20150923_10_0.05_10
% Processing time: 120.904
% Average time per document: 0.119
% Correct mentions: 833
% Total mentions: 1020
% Precision (Micro): 0.8167
% Precision (Macro): 0.8161
+Textual context & 84.6 & - \\
% venv; entity-disambi evaluation --corpus-type=tac_kbp --corpus-tag=eval dataset/tac-kbp-2010/ enwiki_dictionary_20150706.pickle enwiki_alias_db_20150927_tt.pickle ml tac_gbrt_0.05_500_txt_4_0.02_10000.pickle enwiki_entity_vector_500_20150923_10_0.05_10
% Processing time: 114.985
% Average time per document: 0.114
% Correct mentions: 863
% Total mentions: 1020
% Precision (Micro): 0.8461
% Precision (Macro): 0.8457
+Coherence & \textbf{85.5} & - \\
% venv; entity-disambi evaluation --corpus-type=tac_kbp --corpus-tag=eval dataset/tac-kbp-2010/ enwiki_dictionary_20150706.pickle enwiki_alias_db_20150927_tt.pickle ml tac_gbrt_0.05_500_4_0.02_10000.pickle enwiki_entity_vector_500_20150923_10_0.05_10
% Processing time: 89.126
% Average time per document: 0.088
% Correct mentions: 872
% Total mentions: 1020
% Precision (Micro): 0.8549
% Precision (Macro): 0.8546
Two-step & 85.2 & - \\
% venv; entity-disambi evaluation --corpus-type=tac_kbp --corpus-tag=eval dataset/tac-kbp-2010/ enwiki_dictionary_20150706.pickle enwiki_alias_db_20150927_tt.pickle ml tac_two_step_gbrt_0.05_500_4_0.02_10000.pickle enwiki_entity_vector_500_20150923_10_0.05_10
% Processing time: 88.135
% Average time per document: 0.087
% Correct mentions: 866
% Total mentions: 1020
% Precision (Micro): 0.8490
% Precision (Macro): 0.8486
\hline
\end{tabular}
\caption{The results of our feature study.}
\label{tb:feature-study}
\end{table}

We conducted a feature study on our method.
We began with base features, added various features to our system incrementally, and reported their impact on performance.
We then introduced our two-step approach to achieve the final results.

Table \ref{tb:feature-study} shows the results.
Surprisingly, we attained results comparable with those of some state-of-the-art methods on the both datasets by only using base features.
Adding string similarity features slightly further improved performance.

We observed significant improvement when adding textual context features based on our proposed embedding.
Our method outperformed some state-of-the-art methods without using coherence.

Further, coherence based on unambiguous entity mentions and our two-step approach significantly improved performance on the CoNLL dataset.
However, it did not contribute to performance on the TAC 2010 dataset.
This was because of the significant difference in the density of entity mentions between the datasets.
The CoNLL dataset contains approximately 20 entity mentions per document, but the TAC 2010 only contains approximately one mention per document which is unarguably insufficient to model coherence.

\subsubsection{Error Analysis}

We also conducted an error analysis on the CoNLL test set with candidate generation using PPRforNED dataset.
We observed that approximately 48.6\% errors were caused by \textit{metonymy} mentions \cite{Ling2015} (i.e., mentions with more than one plausible annotation).
In particular, our NED method often erred when an incorrect entity was highly popular and exactly matched the mention surface (e.g., ``South Africa'' referring to the entity \textsf{South Africa national rugby union team} rather than the entity \textsf{South Africa}).
This makes sense because our machine learning model uses the popularity statistics of the KB (i.e., prior probability and entity prior), and the string similarity between the title of the entity and the mention surface.
This problem is discussed further in \cite{Ling2015}.

Furthermore, because our method depends on the presence of KB anchors in order to learn entity representation, it arguably fails to learn satisfactory representations of tail entities (i.e., entities rarely referred to by anchors), thus resulting in disambiguation errors.
We discovered that nearly 9.6\% errors were due to referent entities with less than 10 inbound KB anchors, and 4.5\% involved entities with no inbound KB anchor.
These errors might be addressed using KB data other than KB anchors, such as the description of the entities and the KB categories in order to avoid dependence on the KB anchors.
This remains part of our future work.

\section{Related Work}

Early NED methods addressed the problem as a well-studied \textit{word sense disambiguation} problem \cite{Mihalcea2007}.
These methods primarily focused on modeling the similarity of \textit{textual} (\textit{local}) context.
Most recent state-of-the-art methods focus on modeling \textit{coherence} among disambiguated entities in the same document \cite{Cucerzan2007,Milne2008,Hoffart2011,Ratinov2011}.
These approaches have also been called \textit{collective} or \textit{global} approaches in the literature.
% Shen et al. \cite{6823700} recently conducted an extensive survey of NED methods.
% Note that NED is sometimes referred to as \textit{entity linking} (EL).
% However, EL often refers to a broader task because NED usually assumes the presence of an entity referred to by a mention in the KB, whereas EL also requires addressing cases where the entity does not exist in the KB.

Learning the representations of entities for NED has been addressed in past literature.
Guo and Barbosa \cite{Guo2014b} used random walks on KB graphs to construct vector representations of entities and documents to address NED.
Blanco et al. \cite{Blanco2015a} proposed a method to map entities into the word embedding (i.e., Word2vec \cite{Mikolov2013a}) space using entity descriptions in the KB and applied it for NED.
He et al. \cite{he-EtAl:2013:Short} used deep neural networks to compute representations of entities and contexts of mentions directly from the KB.
Similarly, Sun et al. \cite{Sum2015} proposed a method based on deep neural networks to model representations of mentions, contexts of mentions, and entities.
Huang et al. \cite{DBLP:journals/corr/HuangHJ15} also leveraged deep neural networks to learn entity representations such that the consequent pairwise entity relatedness was more suitable than of a standard method (i.e., WLM) for NED.
Further, Hu et al. \cite{hu-EtAl:2015:ACL-IJCNLP} used hierarchical information in the KB to build entity embedding and applied it to model coherence.
Unlike these methods, our proposed approach involves jointly learning vector representations of entities as well as words, hence enabling the accurate computation of the semantic similarity among its items to model both the textual context and coherence.

% Moreover, two recent studies have featured proposals to build an embedding for NED that maps words and entities into the same continuous vector space.
% Blanco et al. \cite{Blanco2015a} proposed an embedding method based on a word embedding (i.e., Word2vec \cite{Mikolov2013,Mikolov2013a}) to model the contexts of NED.
% They adopted logistic regression that learns to distinguish words in the KB description of the entity from other words.
% The learned weights were used as representations of entities.
% Chisholm and Hachey \cite{TACL494} also proposed an embedding method.
% They computed the vector representation of an entity by simply averaging the vectors of words in a corresponding KB description of the entity and context words of anchors that pointed to the entity.
% As in Blanco et al., they used Word2vec as the base word embedding.
% Compared to our model, they naively depended on an existing word embedding, and did not use the link structure of the KB which is known to be highly effective for NED.
% Moreover, Blanco et al. focused only on addressing NED on Web queries, which are different from normal documents.

Moreover, Yaghoobzadeh and Schutze \cite{yaghoobzadeh-schutze:2015:EMNLP} addressed an entity typing task by building an embedding of words and entities on a corpus with annotated entities (i.e., FACC1 \cite{Gabrilovich2013}) using the skip-gram model.
Compared to our method, in addition to the significant difference between their task and NED, their embedding does not incorporate the link graph data of KB, which is known to be highly important for NED.

Furthermore, in the context of \textit{knowledge graph embedding}, another tenor of recent works has been published \cite{AAAI113659,NIPS2013_5028,AAAI159571}.
These methods focus on learning vector representations of entities to primarily address the \textit{link prediction} task that aims to predict a new fact based on existing facts in KB.
Particularly, Wang et al. \cite{wang-EtAl:2014:EMNLP20145} have recently revealed that the joint modeling of the embedding of words and entities can improve performance in several tasks including the link prediction task, which is somewhat analogous to our experimental results.

\section{Conclusions}

In this paper, we proposed an embedding method to jointly map words and entities into the same continuous vector space.
Our method enables us to effectively model both \textit{textual} and \textit{global} contexts.
Further, armed with these context models, our NED method outperforms state-of-the-art NED methods.

In future work, we intend to improve our model by leveraging relevant knowledge, such as relations in a knowledge graph (e.g., Freebase).
We would also like to seek applications of our proposed embedding other than NED.

% The code of the proposed embedding method and the pre-trained vectors used in our experiments will be made publicly available before the conference.

\bibliographystyle{acl2016}
\bibliography{library}
\end{document}